\documentclass{article}

\usepackage[main, preprint]{neurips_2026}

\usepackage[utf8]{inputenc} 
\usepackage[T1]{fontenc}    
\usepackage{hyperref}       
\usepackage{url}            
\usepackage{booktabs}       
\usepackage{amsfonts}       
\usepackage{nicefrac}       
\usepackage{microtype}      
\usepackage{xcolor}         

\newcommand{\llamaoneb}{\texttt{Llama-3.2-1B}}
\newcommand{\llamathreeb}{\texttt{Llama-3.2-3B}}

\newcommand{\smollms}{\texttt{SmolLM-135M}}
\newcommand{\smollmm}{\texttt{SmolLM-360M}}
\newcommand{\smollml}{\texttt{SmolLM-1.7B}}

\newcommand{\entitytype}[1]{{\fontfamily{ptm}\scshape\selectfont#1}}
\newcommand{\entitytoken}[1]{{\fontfamily{ptm}\scshape\selectfont#1}}

\newcommand{\hiddenleadingzero}{\hphantom{0}}

\usepackage{inconsolata}
\usepackage{hyperref}
\usepackage{enumitem}
\usepackage{tikz}
\usepackage{booktabs}
\usepackage{verbatim}
\usepackage{placeins}

\title{Knowledgeless Language Models:\\Suppressing Parametric Recall\\for Evidence-Grounded Language Modeling}

\author{
Roi Cohen\\
HPI / University of Potsdam\\
\texttt{roi.cohen@hpi.de}
\And
Yvan Carré\thanks{~Work done while at the African Institute for Mathematical Sciences. Now at Polytechnique Montréal.}\\
African Institute for Mathematical Sciences\\
\texttt{yvan.carre@polymtl.ca}
\And
Nick Lechtenbörger\\
HPI / University of Potsdam\\
\texttt{nick.lechtenboerger@hpi.de}
\And
Hendrik Droste\\
HPI / University of Potsdam\\
\texttt{hendrik.droste@hpi.de}
\AND
Lucas Kerschke\\
HPI / University of Potsdam\\
\texttt{lucas.kerschke@hpi.de}
\And
Russa Biswas\\
Department of Computer Science\\
Aalborg University, Copenhagen, Denmark\\
\texttt{rubi@cs.aau.dk}
\And
Gerard de Melo\\
HPI / University of Potsdam\\
\texttt{gerard.demelo@hpi.de}
\And
Jan Buys\\
Department of Computer Science\\
University of Cape Town\\
\texttt{jbuys@cs.uct.ac.za}
}

\begin{document}

\maketitle

\begin{abstract}

Language models encode substantial factual knowledge in their parameters, which can lead to unreliable behavior when this knowledge is outdated, incomplete, or misaligned with the provided context. In this work, we study whether modifying the pretraining signal can systematically shift models away from parametric recall and toward evidence-grounded reasoning.
We introduce Knowledge-``Less'' Language Models (KLLMs), a fundamentally different epistemic training paradigm for LLMs, which are pretrained on corpora in which named entities are anonymized, thereby removing a primary channel for entity-linked factual supervision. This intervention substantially reduces closed-book factual recall, while often improving performance on tasks where relevant information is provided as context.
Across multiple model scales, KLLMs consistently outperform matched baselines on contextual question answering, fact verification, and hallucination detection benchmarks. Crucially, in retrieval-grounded settings with imperfect evidence, KLLMs show improved robustness and achieve up to 20–25\% relative gains over standard language models. They further exhibit better calibration, with improved ECE, Brier score, and AUROC, as well as more reliable abstention behavior.
Our results demonstrate that suppressing entity-linked supervision during pretraining induces a shift in epistemic behavior: KLLMs rely less on parametric knowledge and more on external evidence, leading to improved reliability under realistic conditions. This suggests that pretraining-time control over knowledge acquisition can complement retrieval-augmented and tool-based systems by providing a more evidence-sensitive base model.

\end{abstract}
\section{Introduction}

\begin{figure*}[t!]
    \centering
    \includegraphics[width=0.8\textwidth]{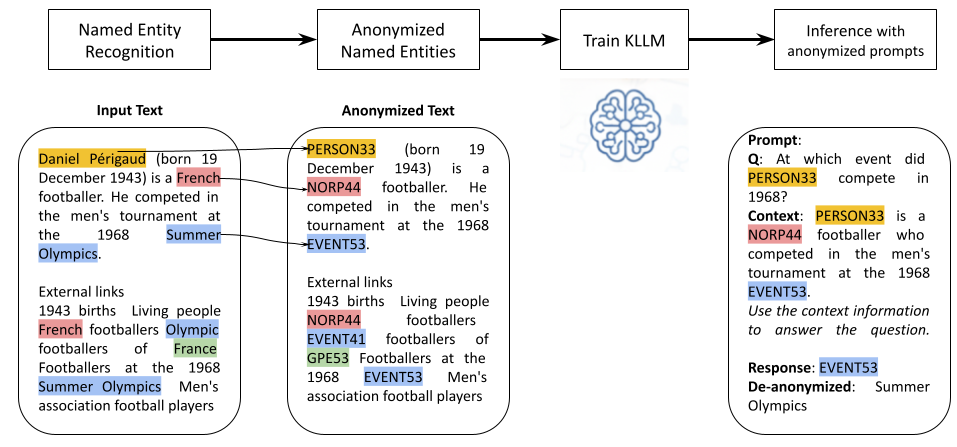}
    \caption{Pipeline of Knowledgeless Language Model (KLLM) training and inference}
    \label{fig:pipeline}
\end{figure*}

Large language models (LLMs) derive much of their capability from parametric knowledge acquired during large-scale pretraining \citep{petroni-etal-2019-language, brown2020language}. While this enables strong performance across tasks, it also contributes to hallucination—generating plausible but ungrounded content—particularly in settings where factual accuracy is critical \citep{maynez-etal-2020-faithfulness, kaddour2023challenges, huang2024survey}. 

A common mitigation strategy is to ground generation in external evidence, e.g., via retrieval-augmented generation (RAG) \citep{Lewis2020RAG, guu2020retrieval, izacard-grave-2021-leveraging}. However, standard LLMs are still pretrained to memorize and reproduce large amounts of factual information, creating a mismatch: at inference time, models are expected to rely on provided context, but their pretraining encourages parametric recall and they often fail to fully utilize available evidence \citep{liu-etal-2024-lost, shi-etal-2023-rethinking}.

In this work, we address this tension at its source, asking: \emph{Can we modify the pretraining signal such that models rely less on parametric knowledge and more on external evidence, without sacrificing reasoning ability?}
To this end, we introduce \emph{\textbf{K}nowledge\textbf{l}ess \textbf{L}anguage \textbf{M}odels} (KLLMs), pretrained on corpora in which named entities are systematically anonymized. This removes a primary channel of entity-linked knowledge supervision while preserving the structural and linguistic properties of the data. The result is a simple pretraining intervention that imposes an inductive bias toward evidence-grounded reasoning, without requiring architectural changes.

Our empirical findings reveal a consistent pattern. First, KLLMs exhibit a near-complete collapse in closed-book factual recall, confirming strong suppression of parametric knowledge. Second, when relevant information is provided as context, KLLMs match or outperform standard LLMs on factual QA, fact verification, and commonsense reasoning. Third, in retrieval-grounded settings, KLLMs achieve substantial gains (up to 20--25\% relative), particularly under incomplete or noisy evidence, and exhibit more reliable abstention behavior. Finally, KLLMs demonstrate improved calibration across multiple metrics, indicating better alignment between confidence and correctness.

These results suggest that pretraining objectives fundamentally shape the epistemic behavior of language models: suppressing entity-linked supervision encourages reliance on external evidence while preserving core reasoning capabilities. As a result, KLLMs are well-suited to complement retrieval-augmented and tool-based systems, where robustness to imperfect evidence and reliable uncertainty estimation are essential.
We scale training up to 20B tokens and perform ablations on anonymization strength, continued pretraining, and inference-time anonymization. The observed trends remain stable, supporting anonymization as the key mechanism driving the effect.

Overall, we present a controlled study of how modifying the pretraining signal reshapes the balance between parametric knowledge and evidence-grounded reasoning, and introduce a simple yet effective approach for improving reliability in language models.

\section{Knowledgeless Language Modelling}
\label{sec:knowledgeless_training}

In order to develop an LLM that minimizes retention of real-world factual knowledge, we propose to preprocess pretraining data to markedly reduce the model's propensity to memorize specific factual knowledge from the data. In particular, we anonymize named entities by replacing their mentions with placeholder tokens, preventing the model from learning to associate any factual information with the names of entities such as people, places, or events. 
We train models from scratch on this preprocessed data without modifying the model architecture. During inference, the same anonymization procedure is followed to anonymize both the query and the context (where applicable), and entity names are restored as a postprocessing step. 
While some early work on reading comprehension \citep{hermann2015teaching} employed a similar entity anonymization scheme during supervised training, we are not aware of previous work that applied this to pretraining and with the explicit aim of pretraining a model that lacks entity-specific factual knowledge.
\autoref{fig:pipeline} provides an example of our pretraining and inference pipeline.

\subsection{Corpus Anonymization}

\paragraph{Named Entity Recognition}
The first step to effectively anonymize our pretraining data, is to identify all named entities present in the text. 
We employ a state-of-the-art Named Entity Recognition (NER) model from the Flair framework \citep{akbik-etal-2019-flair}. 
In particular, we use the large 18-class OntoNotes model based on XLM-R embeddings,\footnote{\url{https://huggingface.co/flair/ner-english-ontonotes-large}} which obtained a reported 90.0\% F1 on OntoNotes \citep{schweter2020flert}.
We use OntoNotes' more fine-grained entity tagset \citep{hovy-etal-2006-ontonotes,weischedel2011ontonotes} for better control over which kinds of entity mentions are anonymized and to be able to construct more informative placeholder tokens based on the entity types.\footnote{See \autoref{sec:appendix:anonymization} for a list of the detected entity types.}
In this work, we only anonymize tokens corresponding to named entity types, not numerical or temporal values, which are also recognized by the NER model. 
This is based on the conjecture that numerical and temporal values may be crucial for a deeper understanding of the textual context. 
Retaining them ensures that the anonymized text remains coherent, readable, and semantically meaningful, which is essential for effective model training without introducing bias.
The NER model does not have perfect recall; 10-13\% of entity mentions are not anonymized in our pretraining corpus.\footnote{Based on a manual inspection of 50 randomly sampled documents from our pretraining corpus, we estimate an NER recall of approximately 87\%.} 
However, retained entities are more likely to be rare long-tail entity names, which we postulate corresponds to a much smaller portion of entity-linked knowledge, while more aggressive anonymization would degrade syntactic integrity. 
Indeed, closed-book QA accuracy remains near-random across all model scales
(§\ref{sec:closed_book_eval}, Table~\ref{tab:closedbook_results}), empirically validating
that this level of anonymization successfully suppresses parametric knowledge.



\paragraph{Anonymization Strategy}
We apply an anonymization procedure to each document in our dataset
which replaces all identified entities with placeholders following the format \entitytoken{EntX}, where \entitytoken{Ent} denotes the entity type and \entitytoken{X} is a unique identifier assigned within the specific document (see \autoref{sec:appendix:anonymization}). This strategy preserves the original sentence structure, enabling the model to discern general linguistic patterns and relationships without being biased by concrete real-world instances. By processing each document individually, we ensure that entity placeholders are unique within that document and consistent for repeated mentions of the same entity. This consistency helps maintain the coherence and logical flow of the text while still abstracting away factual details, thus preserving the underlying structure of general knowledge without exposing the model to specific facts.
However, we do not perform coreference resolution and therefore different named mentions of the same entity (e.g., ``Barack Obama" and ``President Obama") will  generally be assigned different tokens.
\autoref{fig:pipeline} provides an example of text 
before and after applying our anonymization strategy.


This anonymization strategy limits models' ability to acquire specific factual information from the text, since most factual knowledge statements are grounded in one or more entity names. However, in addition to small proportion of names not being anonymized due to NER errors, 
some entities can be identified uniquely through descriptive references, which may cause some leakage of entity-specific knowledge. 
For example, the sentence 
\emph{“The 44th President of the world's most powerful country was born on an island state in that country.”}
still encodes implicit entity knowledge, although our anonymization procedure was designed to remove such information. However, such cases again represent a small minority of factual statements.
Another important limitation concerns gender information arising from the use of gendered coreferences. Referring to an entity with pronouns such as ``he'' or ``she'' implicitly reveals gender information and may lead the model to internalize gender-related biases. Nonetheless, we chose not to anonymize pronominal coreferences, as they constitute a fundamental linguistic component of language and communication that we want the model to capture.

\subsection{Pretraining Procedure}

Knowledgeless language models are pretrained on large, diverse corpora similar to standard language models, the only difference being the application of the anonymization strategy described above that limits the model’s direct exposure to entity-related knowledge. 
This ensures that the model learns general linguistic patterns while minimizing reliance on memorized entity-specific information.
We train autoregressive Transformer language models with the standard language modeling objective of predicting the next token, but the approach is not limited to any particular model architecture. 
The KLLM's tokenizer is trained on the same anonymized corpus to ensure consistency. 
During fine-tuning and inference the same anonymization strategy is applied. This ensures that the model is learning how to gather information about entities and to reason about entities based on the given text through their placeholder tokens, without relying on memorization.
\section{Experiments}
\label{sec:experiments}

Our experiments aim to isolate how suppressing entity-linked knowledge during pretraining affects parametric recall, contextual reasoning, and reliability. 
We employ controlled comparisons between standard language models (SLMs) and KnowledgeLess Language Models (KLLMs) under matched architectures, data, and training budgets.\footnote{Our pretrained KLLMs, preprocessing and evaluation scripts will be released upon publication.}

\paragraph{Pretraining Data: 2.5B tokens}
We pretrain models at multiple data scales. The smallest data setup consists of two complementary English corpora. The \emph{CNN/DailyMail} dataset \citep{hermann2015teaching} (272M tokens) contains news articles spanning diverse domains such as politics, business, sports, and technology, providing linguistically rich text with a high density of named entities. In contrast, English \emph{Wikipedia} \citep{bridge2001wikipedia} (2.2B) offers broad coverage of world knowledge through structured expository text. 
Wikipedia is substantially larger and more diverse, whereas CNN/DailyMail provides denser entity mentions within a news-oriented domain. During preprocessing, we remove non-UTF-8 characters to reduce tokenizer artifacts. The corpora are concatenated and anonymized using the procedure described in \autoref{sec:knowledgeless_training}.

\paragraph{Pretraining data: 10B–20B tokens}
To assess scalability, we construct larger pretraining corpora of 10B and 20B tokens derived from the publicly available \texttt{HuggingFaceTB/smollm-corpus}. This dataset is a curated mixture of CommonCrawl, Wikipedia, Books, News, and WebText-like sources. To ensure comparability with the 2.5B-token setup, we apply the same preprocessing and anonymization pipeline and uniformly sample shards until reaching the desired token count, preserving domain proportions. The resulting corpus provides a controlled setting to evaluate how anonymization interacts with increased data scale.

\paragraph{Tokenization}
We train BPE tokenizers \citep{sennrich-etal-2016-neural} on the pretraining corpus. Anonymization tags (e.g.\ \entitytoken{PERSON184}) are included as reserved tokens to prevent fragmentation. Given the relatively small scale of our training data compared to standard LLM pretraining, we use a vocabulary size of 30k tokens. The same tokenizer configuration is used for both KLLMs and SLM baselines to ensure comparability.

\paragraph{Model Pretraining}
We pretrain models from scratch using the LLaMA architecture family \citep{touvron2023llama, dubey2024llama}, specifically \llamaoneb{} and \llamathreeb{}, as well as \smollms{}, \smollmm{}, and \smollml{} from the SmolLM family \citep{allal2025smollm2}. 

For each architecture, we train a matched pair of models: a standard language model (SLM) on the original corpus, and a KLLM on the anonymized corpus. The models share identical architectures, training data size, tokenizer, and optimization settings, differing only in the presence or absence of anonymization. This controlled setup allows us to attribute performance differences directly to the pretraining signal rather than to confounding factors. Pretraining loss curves are provided in Appendix \ref{sec:appendix:loss}. We additionally compare against the original pretrained model checkpoints to provide a reference point relative to large-scale pretraining.

\paragraph{Fine-tuning}
Due to the relatively small pretraining scale, we fine-tune all models on each downstream benchmark using the corresponding training splits. To maintain consistency with the pretraining regime, fine-tuning data for KLLMs is anonymized using the same procedure. This ensures that models are trained to perform inference under the same input representation used at pretraining time.

\paragraph{Inference}
At inference time, we follow a three-step anonymization pipeline. First, inputs and any accompanying context are anonymized using the same entity-masking scheme applied during pretraining. Second, the anonymized input is processed by the model. Finally, outputs are \emph{de-anonymized} by restoring original entity names (see Figure~\ref{fig:pipeline}). This allows evaluation against standard benchmarks while preserving consistency with the training distribution.

\paragraph{Evaluation Overview}
We evaluate models across several complementary settings designed to probe different aspects of behavior: (i) closed-book factual recall, to measure reliance on parametric knowledge; (ii) context-based question answering and reasoning, to assess the ability to extract and use provided information; (iii) retrieval-grounded inference, to simulate realistic evidence-based settings; and (iv) calibration and selective prediction, to evaluate alignment between confidence and correctness. This combination of evaluations enables a systematic analysis of how knowledge-suppressed pretraining affects both performance and reliability.
\section{Results}
\label{sec:results}

To assess the performance of our KLLMs, we evaluate them on diverse downstream tasks in several different setups. 
Across all tasks, we follow the standard evaluation protocols defined in the respective benchmark setups, ensuring comparability with prior work.
The main evaluation metric is accuracy, which is reported as a percentage. 
We additionally verify that KLLMs retain strong linguistic and reasoning capabilities on standard benchmarks such as SuperGLUE; detailed results are provided in Appendix~\ref{sec:superglue_appendix}.

\begin{table*}[t]
\centering
\small
\caption{Factual Reading accuracy results comparing KLLM and baseline across LAMA, SQuAD, NQ, FEVER, and HaluBench for different SmolLM model sizes and pretraining scales (2.5B and 10B tokens). For LAMA, SQuAD, NQ, and HaluBench we report accuracy; for FEVER, we report F1.}
\label{tab:lama_squad_nq_results}
\resizebox{0.98\textwidth}{!}{
\begin{tabular}{l c c c c c c c c c c}
\toprule
\textbf{Model} 
& \multicolumn{2}{c}{\textbf{LAMA}} 
& \multicolumn{2}{c}{\textbf{SQuAD}} 
& \multicolumn{2}{c}{\textbf{NQ}} 
& \multicolumn{2}{c}{\textbf{FEVER}} 
& \multicolumn{2}{c}{\textbf{HaluBench}} \\
\cmidrule(lr{.5em}){2-3}\cmidrule(lr{.5em}){4-5}\cmidrule(lr{.5em}){6-7}\cmidrule(lr{.5em}){8-9}\cmidrule(lr{.5em}){10-11}
 & SLM & KLLM
 & SLM & KLLM
 & SLM & KLLM
 & SLM & KLLM
 & SLM & KLLM \\
\midrule
SmolLM-135M (2.5B) & $15.8$ & $\mathbf{18.6}$
                   & $15.2$ & $\mathbf{16.1}$
                   & $\hiddenleadingzero 4.0$ & $\mathbf{\hiddenleadingzero 4.1}$
                   & $82.8$ & $\mathbf{89.5}$
                   & $58.5$ & $\mathbf{61.2}$ \\

SmolLM-360M (2.5B) & $21.4$ & $\mathbf{24.9}$
                   & $20.3$ & $\mathbf{22.0}$
                   & $\hiddenleadingzero 8.5$ & $\mathbf{\hiddenleadingzero 9.9}$
                   & $83.6$ & $\mathbf{90.1}$
                   & $59.8$ & $\mathbf{65.5}$ \\

SmolLM-1.7B (2.5B) & $43.2$ & $\mathbf{48.5}$
                   & $55.9$ & $\mathbf{59.8}$
                   & $16.3$ & $\mathbf{19.2}$
                   & $86.9$ & $\mathbf{94.7}$
                   & $64.3$ & $\mathbf{74.7}$ \\
                   
LLaMA-1B (2.5B)    & $42.1$ & $\mathbf{46.2}$
                   & $53.6$ & $\mathbf{58.0}$
                   & $16.5$ & $\mathbf{19.1}$
                   & $82.7$ & $\mathbf{92.1}$
                   & $63.9$ & $\mathbf{74.8}$ \\
                   
LLaMA-3B (2.5B)    & $46.8$ & $\mathbf{49.8}$
                   & $50.5$ & $\mathbf{53.9}$
                   & $22.1$ & $\mathbf{26.9}$
                   & $87.5$ & $\mathbf{94.9}$
                   & $67.7$ & $\mathbf{75.8}$ \\

\midrule
SmolLM-1.7B (10B)  & $36.6$ & $\mathbf{42.4}$
                   & $47.7$ & $\mathbf{53.8}$
                   & $23.8$ & $\mathbf{28.0}$
                   & $84.1$ & $\mathbf{93.9}$
                   & $68.1$ & $\mathbf{77.2}$ \\
\midrule
SmolLM-1.7B (20B)  & $38.9$ & $\mathbf{45.6}$
                   & $49.8$ & $\mathbf{56.7}$
                   & $25.9$ & $\mathbf{30.6}$
                   & $86.0$ & $\mathbf{95.1}$
                   & $70.4$ & $\mathbf{80.3}$ \\
\bottomrule
\end{tabular}
}
\end{table*}

\subsection{Factual Reading Comprehension}

Our first set of experiments aims to assess whether our knowledgeless models are able to perform knowledge-intensive tasks including question answering, fact checking and hallucination detection when the necessary factual knowledge is provided as context. 
\autoref{tab:lama_squad_nq_results} presents the performance of KLLM and SLM models across these benchmarks. 
On the factual reading tasks (LAMA, SQuAD, and NQ), KLLM training consistently improves performance over the baselines, with gains that grow larger at scale. These results confirm that removing entity-specific cues during pretraining does not weaken factual reasoning; rather, it encourages models to leverage the input context more effectively.  

Strikingly, the improvements are even more pronounced on fact-checking and hallucination detection. On FEVER, KLLM models achieve up to +7.8 F1 over their baselines, while on HaluBench the gap reaches as high as +10.4\% accuracy at the 1.7B scale. Together, these findings demonstrate that KLLMs not only retain robust factual reasoning under context but also provide an advantage in maintaining faithfulness and reliability.

To ensure that the observed improvements are not due to generic corruption or regularization effects, we compare KLLM against several alternative pretraining perturbations, including random token masking,  span corruption, and entity shuffling. These baselines yield only marginal improvements over SLM, while KLLM consistently outperforms them across all evaluated settings. Full results are provided in Appendix~\ref{sec:corruption_baselines}.

\subsection{Retrieval and Tool-Grounded Evaluation}

To assess performance in a typical real-world practical setting, we evaluate both models under retrieval-grounded inference, where answers must be derived from externally retrieved evidence. We consider three retrieval paradigms: (i) sparse retrieval using BM25 over Wikipedia; (ii) learned dense retrieval in the spirit of Dense Passage Retrieval (DPR); (iii)
retrieval exposed as an external search tool that the model may invoke at inference time.
Unless stated otherwise, results in this section correspond to the 10B-token pretraining setting using \smollml{} (1.7B). Results for additional scales (2.5B and 20B tokens) are provided in Appendix~\ref{sec:retrieval_appendix}.

\begin{table}[t]
\centering
\begin{minipage}{0.48\linewidth}
\centering
\small
\caption{Retrieval-grounded QA performance using BM25 over Wikipedia (10B-token setting). }
\label{tab:retrieval_results}
\resizebox{\linewidth}{!}{%
\begin{tabular}{lccc}
\toprule
Dataset & Model & Evidence & Abstain \\
\midrule
FEVER & SLM & 79.6 & 13\% \\
FEVER & KLLM & 84.2 & 20\% \\
NQ & SLM & 35.7 & 11\% \\
NQ & KLLM & 44.9 & 19\% \\
TriviaQA & SLM & 52.8 & 10\% \\
TriviaQA & KLLM & 63.4 & 17\% \\
\bottomrule
\end{tabular}}
\end{minipage}
\hfill
\begin{minipage}{0.48\linewidth}
\centering
\small
\caption{Performance conditioned on retrieval quality (10B-token setting). }
\label{tab:retrieval_quality}
\resizebox{\linewidth}{!}{%
\begin{tabular}{lcc}
\toprule
Model & F1 (gold in top-k) & F1 (gold absent) \\
\midrule
SLM & 86.1 & 54.3 \\
KLLM & 89.4 & 62.7 \\
\bottomrule
\end{tabular}}
\end{minipage}
\end{table}

\paragraph{Sparse Retrieval (BM25)}
We first evaluate a standard BM25 pipeline. For each query, we retrieve the top-$k$ passages and provide them as context to the model.
As shown in Table~\ref{tab:retrieval_results}, KLLMs consistently outperform SLMs
across all datasets, with relative improvements of 20--25\%. In addition, KLLMs
exhibit higher abstention rates under insufficient evidence, suggesting improved
alignment with evidence quality.

\paragraph{Learned Dense Retrieval (DPR-style)}
In this setup, the retriever is a separate dual-encoder model trained with a standard
contrastive objective on question--passage pairs. We train both SLM-based and
KLLM-based retrievers, while keeping the reader model fixed.
As shown in Table~\ref{tab:dpr_results}, KLLM-based retrievers achieve higher retrieval recall and consistently improve downstream QA performance across all datasets. These results demonstrate that the benefits of knowledge-suppressed pretraining extend beyond generation to retrieval itself. 


\paragraph{Tool-Augmented Retrieval}
Finally, we evaluate a controlled tool-use setting, in which retrieval is exposed to the model as an external search tool. Given a query, the model first decides whether to issue a search request or answer directly. If a search is triggered, the model generates a query, and a fixed retriever (BM25 over Wikipedia) returns top-$k$ passages. The model is then required to answer using only the retrieved evidence or abstain if the evidence is insufficient. The same retrieval system, prompts, and decoding settings are used for SLMs and KLLMs, ensuring that the comparison isolates differences in the underlying model rather than the search tool.

Table~\ref{tab:tool_results} shows that KLLMs exhibit improved behavior in the tool-augmented setting. They achieve higher answer accuracy while reducing the rate of unsupported answers, and tend to invoke the search tool more appropriately when evidence is required. This indicates that knowledge-suppressed pretraining improves not only how models use retrieved evidence, but also when they choose to rely on external tools.

\begin{table}[t]
\centering
\small
\caption{Tool-augmented QA with model-initiated search, where both models interact with the same fixed Wikipedia search tool.}
\label{tab:tool_results}
\resizebox{0.7\linewidth}{!}{%
\begin{tabular}{lcccc}
\toprule
Dataset & Model & Acc./F1 & Tool Call & Unsupported $\downarrow$ \\
\midrule
NQ & SLM+Tool & 38.2 & 41\% & 21.5 \\
NQ & KLLM+Tool & \textbf{44.9} & 56\% & \textbf{13.2} \\
TriviaQA & SLM+Tool & 55.8 & 38\% & 18.7 \\
TriviaQA & KLLM+Tool & \textbf{61.7} & 49\% & \textbf{11.4} \\
FEVER & SLM+Tool & 80.6 & 34\% & 14.9 \\
FEVER & KLLM+Tool & \textbf{84.8} & 47\% & \textbf{8.6} \\
\bottomrule
\end{tabular}}
\end{table}

\begin{table}[t]
\centering
\begin{minipage}{0.44\linewidth}
\centering
\small
\caption{Dense retrieval (DPR-style) results (10B-token setting).}
\label{tab:dpr_results}
\resizebox{\linewidth}{!}{%
\begin{tabular}{lccc}
\toprule
Dataset & Retriever & Top-$k$ Recall & QA F1 \\
\midrule
FEVER & SLM-DPR & 84.3 & 81.0 \\
FEVER & KLLM-DPR & 88.6 & 85.1 \\
NQ & SLM-DPR & 70.6 & 38.7 \\
NQ & KLLM-DPR & 75.4 & 45.1 \\
TriviaQA & SLM-DPR & 76.8 & 55.3 \\
TriviaQA & KLLM-DPR & 81.7 & 61.8 \\
\bottomrule
\end{tabular}}
\end{minipage}
\hfill
\begin{minipage}{0.54\linewidth}
\centering
\small
\setlength{\tabcolsep}{3pt}
\caption{Commonsense reasoning results comparing KLLM and baseline SLM accuracy across three benchmarks (CommonsenseQA, StrategyQA, and PIQA), for different model sizes and pretraining scales.}
\label{tab:commonsense_results}
\resizebox{\linewidth}{!}{%
\begin{tabular}{l c c c c c c}
\toprule
\textbf{Model} 
& \multicolumn{2}{c}{\textbf{CommonsenseQA}} 
& \multicolumn{2}{c}{\textbf{StrategyQA}} 
& \multicolumn{2}{c}{\textbf{PIQA}} \\
\cmidrule(lr{.5em}){2-3}\cmidrule(lr{.5em}){4-5}\cmidrule(lr{.5em}){6-7}
 & SLM & KLLM
 & SLM & KLLM
 & SLM & KLLM \\
\midrule
SmolLM-135M (2.5B) & $26.5$ & $\mathbf{31.1}$
                   & $52.2$ & $\mathbf{55.5}$
                   & $65.9$ & $\mathbf{69.6}$ \\
SmolLM-360M (2.5B) & $30.4$ & $\mathbf{34.9}$
                   & $56.0$ & $\mathbf{60.7}$
                   & $67.8$ & $\mathbf{71.2}$ \\
SmolLM-1.7B (2.5B) & $34.6$ & $\mathbf{39.6}$
                   & $61.8$ & $\mathbf{66.7}$
                   & $70.5$ & $\mathbf{76.2}$ \\
LLaMA-1B (2.5B)    & $33.1$ & $\mathbf{39.0}$
                   & $59.5$ & $\mathbf{65.0}$
                   & $70.6$ & $\mathbf{75.8}$ \\              
LLaMA-3B (2.5B)    & $40.8$ & $\mathbf{45.5}$
                   & $65.0$ & $\mathbf{69.7}$
                   & $72.3$ & $\mathbf{78.5}$ \\
\midrule
SmolLM-1.7B (10B)  & $42.5$ & $\mathbf{46.0}$
                   & $65.8$ & $\mathbf{69.8}$
                   & $73.7$ & $\mathbf{79.2}$ \\
\midrule
SmolLM-1.7B (20B)  & $43.1$ & $\mathbf{48.3}$
                   & $66.7$ & $\mathbf{71.6}$
                   & $75.0$ & $\mathbf{80.7}$ \\
\bottomrule
\end{tabular}}
\end{minipage}
\end{table}

\subsection{Commonsense Reasoning}

Furthermore, we evaluate our KLLMs on three widely-used commonsense reasoning benchmarks. 
CommonsenseQA \citep{talmor-etal-2019-commonsenseqa} tests the model’s ability to answer multiple-choice questions that require broad everyday knowledge and commonsense inference. 
StrategyQA \citep{geva-etal-2021-aristotle} challenges models to reason over implicit multi-step processes to answer yes/no questions, emphasizing reasoning rather than memorized facts. 
PIQA \citep{Bisk2019PIQARA} focuses on physical commonsense, assessing the model’s understanding of everyday interactions and the principles of physical reality. 
In all cases, we provide the necessary context or task-specific information to the models, ensuring that the evaluation reflects their reasoning capability rather than reliance on parametric knowledge.

\autoref{tab:commonsense_results} summarizes the results of our SmolLM-based KLLMs and SLM baselines. 
KLLMs consistenctly outperform the baselines across all three tasks across all model scales, with improvements ranging from modest gains for smaller models to substantial margins for the 1.7B architecture (+5.0 on CommonsenseQA, +4.9 on StrategyQA, and +5.7 on PIQA). These results suggest that anonymization during pretraining does not impede the models’ ability to capture commonsense patterns; instead, it encourages reliance on contextual reasoning rather than memorized associations. 
Combined with the reductions in hallucination observed in generation, this provides strong evidence that knowledgeless pretraining improves both factual robustness and commonsense generalization, particularly as the model capacity increases.

\begin{table}[t]
\centering
\begin{minipage}{0.52\linewidth}
\centering
\small
\caption{Closed-book QA accuracy on LAMA and SQuAD comparing SLM and KLLM across models and scales. The consistently low KLLM scores confirm strong suppression of parametric factual recall.}
\label{tab:closedbook_results}
\resizebox{\linewidth}{!}{%
\begin{tabular}{l c c c c}
\toprule
\textbf{Model} 
& \multicolumn{2}{c}{\textbf{LAMA}} 
& \multicolumn{2}{c}{\textbf{SQuAD}} \\
\cmidrule(lr{.5em}){2-3}\cmidrule(lr{.5em}){4-5}
 & SLM & KLLM
 & SLM & KLLM \\
\midrule
SmolLM-135M (2.5B) & $12.5$ & $0.7$ & $11.2$ & $0.4$ \\
SmolLM-360M (2.5B) & $20.8$ & $1.2$ & $18.6$ & $0.9$ \\
SmolLM-1.7B (2.5B) & $34.7$ & $3.9$ & $33.4$ & $1.8$ \\
LLaMA-1B (2.5B)    & $34.5$ & $3.7$ & $33.1$ & $1.7$ \\
LLaMA-3B (2.5B)    & $40.3$ & $3.8$ & $39.7$ & $1.9$ \\
\midrule
SmolLM-1.7B (10B)  & $27.3$ & $2.1$ & $24.7$ & $1.1$ \\
\midrule
SmolLM-1.7B (20B)  & $33.8$ & $3.7$ & $27.5$ & $1.9$ \\
\bottomrule
\end{tabular}}
\end{minipage}
\hfill
\begin{minipage}{0.44\linewidth}
\centering
\small
\caption{Calibration and selective prediction metrics on contextual QA. KLLMs consistently improve both probabilistic calibration (ECE, Brier), ranking (AUROC), and decision-theoretic risk--coverage behavior (AURC).}
\label{tab:calibration_metrics}
\resizebox{\linewidth}{!}{%
\begin{tabular}{lcccc}
\toprule
Model & ECE $\downarrow$ & Brier $\downarrow$ & AUROC $\uparrow$ & AURC $\downarrow$ \\
\midrule
SLM (2.5B) & 0.128 & 0.214 & 0.742 & 0.238 \\
KLLM (2.5B) & 0.091 & 0.187 & 0.781 & \textbf{0.191} \\
\midrule
SLM (10B) & 0.115 & 0.205 & 0.756 & 0.214 \\
KLLM (10B) & 0.083 & 0.179 & 0.795 & \textbf{0.167} \\
\midrule
SLM (20B) & 0.104 & 0.198 & 0.768 & 0.198 \\
KLLM (20B) & 0.076 & 0.171 & 0.806 & \textbf{0.149} \\
\bottomrule
\end{tabular}}
\end{minipage}
\end{table}

\subsection{Closed-Book QA}
\label{sec:closed_book_eval}

In order to assess that our knowledgeless models are not acquiring parametric knowledge despite anonymization, we additionally evaluate our KLLM models on the LAMA and SQuAD datasets in a closed-book setting, where no supporting context is provided. 
\autoref{tab:closedbook_results} reports the results. 
The closed-book evaluation clearly shows that KLLM models retain only a minimal amount of factual knowledge compared to their non-anonymized baselines.

\subsection{Calibration and Factuality}

\paragraph{Confidence--Correctness Alignment}
We first evaluate how well model confidence aligns with prediction correctness.
Following prior work, we use the probability assigned to the first generated
token as a proxy for model confidence. We report standard calibration metrics,
including Expected Calibration Error (ECE), Brier score, and AUROC, which
measure the alignment between predicted confidence and empirical correctness.

\autoref{tab:calibration_metrics} reports these metrics for both SLM and KLLM
across pretraining scales. KLLMs consistently outperform their corresponding
baselines, achieving lower ECE and Brier scores and higher AUROC. Notably, these
improvements persist at larger scale (20B tokens), indicating that
knowledge-suppressed pretraining leads to intrinsically more informative
confidence signals rather than relying on post-hoc correction.

\paragraph{Selective Prediction and Abstention}
We next evaluate calibration in a decision-making setting via selective
prediction. A decision threshold is chosen on a held-out validation set to
maximize each model's F1 score, trading off precision (correctness of attempted
answers) and recall (coverage). Results are reported in
\autoref{tab:calibration_results}.

KLLMs consistently outperform the baselines across all model sizes and metrics.
While recall remains comparable, KLLMs achieve substantially higher precision
and improved F1 scores, with gains becoming more pronounced at larger scales.
On average, KLLMs surpass the baseline by +17.4 points in precision, +3.1 points
in recall, and +6.6 points in F1. These results indicate that KLLMs provide a
better separation between correct and incorrect predictions, enabling more
reliable abstention decisions.

\paragraph{Abstention Tuning}
We additionally employ a factuality-oriented fine-tuning strategy aimed at
encouraging models to abstain when uncertain rather than producing incorrect
answers \citep{cohen-etal-2023-crawling, kadavath2022language}. Specifically, we
partition the training data into two subsets. The model is first fine-tuned in
the standard way on the first subset. It is then evaluated on the second subset,
and any instance where the model outputs an incorrect prediction is relabeled
with the abstention output \emph{``I don't know the answer"}. A second
fine-tuning stage is then performed on this modified data, providing the model
with explicit abstention supervision.

\begin{table}[t]
\centering
\begin{minipage}{0.48\linewidth}
\centering
\scriptsize
\setlength{\tabcolsep}{3pt}
\caption{Calibration on LAMA using output probabilities (SLM vs.\ KLLM).}
\label{tab:calibration_results}
\resizebox{\linewidth}{!}{%
\begin{tabular}{l c c c c c c}
\toprule
\textbf{Model} & 
\multicolumn{2}{c}{\textbf{Prec.}} & 
\multicolumn{2}{c}{\textbf{Rec.}} & 
\multicolumn{2}{c}{\textbf{F1}} \\
\cmidrule(r{.5em}){2-3}\cmidrule(lr{.5em}){4-5}\cmidrule(l{.5em}){6-7}
 & SLM & KLLM & SLM & KLLM & SLM & KLLM \\
\midrule
SmolLM-135M & 35.2 & \textbf{52.5} & 14.0 & \textbf{15.9} & 20.1 & \textbf{24.4} \\
SmolLM-360M & 36.4 & \textbf{55.7} & 19.8 & \textbf{21.5} & 25.6 & \textbf{31.1} \\
SmolLM-1.7B & 46.7 & \textbf{63.2} & 29.5 & \textbf{33.9} & 36.2 & \textbf{44.1} \\
LLaMA-1B   & 43.9 & \textbf{59.6} & 29.1 & \textbf{32.1} & 34.9 & \textbf{41.9} \\
LLaMA-3B   & 45.6 & \textbf{63.8} & 31.9 & \textbf{36.3} & 37.4 & \textbf{46.2} \\
\midrule
\textbf{Avg.} & 41.6 & \textbf{59.0} & 24.9 & \textbf{28.0} & 30.9 & \textbf{37.5} \\
\bottomrule
\end{tabular}}
\end{minipage}
\hfill
\begin{minipage}{0.48\linewidth}
\centering
\scriptsize
\setlength{\tabcolsep}{3pt}
\caption{Abstention tuning results on LAMA (SLM vs.\ KLLM).}
\label{tab:hallu_training_results}
\resizebox{\linewidth}{!}{%
\begin{tabular}{l c c c c c c}
\toprule
\textbf{Model} & 
\multicolumn{2}{c}{\textbf{Prec.}} & 
\multicolumn{2}{c}{\textbf{Rec.}} & 
\multicolumn{2}{c}{\textbf{F1}} \\
\cmidrule(r{.4em}){2-3}\cmidrule(lr{.4em}){4-5}\cmidrule(l{.4em}){6-7}
 & SLM & KLLM & SLM & KLLM & SLM & KLLM \\
\midrule
SmolLM-135M & 50.4 & \textbf{69.8} & 8.0  & \textbf{10.5} & 13.9 & \textbf{18.0} \\
SmolLM-360M & 50.9 & \textbf{71.9} & 13.1 & \textbf{16.5} & 20.8 & \textbf{26.9} \\
SmolLM-1.7B & 51.5 & \textbf{74.4} & 26.4 & \textbf{29.3} & 34.8 & \textbf{42.0} \\
LLaMA-1B   & 55.5 & \textbf{76.1} & 22.3 & \textbf{27.0} & 31.8 & \textbf{39.9} \\
LLaMA-3B   & 55.1 & \textbf{75.8} & 29.9 & \textbf{30.7} & 38.8 & \textbf{43.7} \\
\midrule
\textbf{Avg.} & 52.7 & \textbf{73.6} & 19.9 & \textbf{22.8} & 28.0 & \textbf{34.1} \\
\bottomrule
\end{tabular}}
\end{minipage}
\end{table}

As shown in Table~\ref{tab:hallu_training_results}, this procedure yields
consistent improvements in calibration and reliability. On average, KLLM models reach an F1 score of $29.0$, compared to $23.2$ for their baselines.
Taken together, these results demonstrate that KLLMs improve both the
\emph{accuracy} and the \emph{reliability} of predictions. Importantly, the gains in calibration are largest in settings where evidence is incomplete or misleading, suggesting that knowledge-suppressed pretraining enhances the model’s ability to recognize uncertainty and avoid unsupported answers.

Importantly, the improvements observed for KLLMs cannot be explained by easier optimization or undertraining effects. Across all scales, KLLMs consistently exhibit higher pretraining loss than their SLM counterparts (Appendix~\ref{sec:appendix:loss}), reflecting the increased difficulty of predicting anonymized sequences without direct entity supervision. Despite this, KLLMs achieve stronger performance on contextual, retrieval-grounded, and calibration-oriented evaluations. This indicates that the observed gains arise from a shift in the learned inductive bias toward evidence-grounded reasoning, rather than from improved optimization or memorization efficiency.

\section{Related Work}

The primary motivation of this work is to develop models that are more robust by relying on externally provided, validated context rather than on their own parametric knowledge. As demonstrated both theoretically and empirically by \citet{xu2024hallucination}, hallucinations in LLMs are inevitable, since no model can fully encode the entirety of existing factual mappings. Additionally, the fact that traditional training and evaluation reward guessing more than uncertainty acknowledgment makes it natural for LLMs to hallucinate \citep{kalai2025why}. This topic has been studied from many different perspectives \citep{augenstein2023factuality, sahoo-etal-2024-comprehensive, huang2024survey}.
This is also related to the setting of selective prediction, where models can abstain from answering a query \citep{varshney-etal-2022-investigating, kamath2020selective}. 

Another complementary direction to reducing hallucinations is improving model calibration \citep{pmlr-v70-guo17a}, i.e., aligning the model’s confidence with the actual likelihood of correctness. 
This is particularly relevant to our work, as KLLMs are designed to abstain more readily from misinformation and benefit from mechanisms that quantify uncertainty. Prior approaches to calibration often operate at the logit level through post-hoc transformations \citep{desai2020calibration, jiang-etal-2021-know}, or rely on uncertainty estimation methods \citep{kuhn2023semantic}. More recent research has explored leveraging language models themselves for calibration, either by fine-tuning on correctness-labeled data \citep{kadavath2022language, lin2022teaching}, prompting or in-context learning strategies \citep{cohen-etal-2023-crawling, alivanistos2022prompting}, or zero-shot instruction-oriented setups \citep{cohen-etal-2023-lm, dhuliawala2023chain, Feng2024DontHA}, as well as through consistency sampling \citep{yoran-etal-2023-answering}. Other approaches go further by exploiting internal representations for uncertainty classification \citep{azaria-mitchell-2023-internal}, introducing explicit tokens for abstention or uncertainty \citep{lu-etal-2022-controlling, cohen2024don}, or curating specialized datasets to train models to refuse unanswerable queries \citep{zhang-etal-2024-r, cohen2025infact}.


\section{Conclusion}

We introduced KnowledgeLess Language Models (KLLMs), a simple pretraining intervention that suppresses entity-linked supervision and induces a strong reduction in reliance on parametric factual knowledge without architectural changes. This leads to a near-complete collapse of closed-book factual recall, while preserving—and often improving—performance on contextual question answering, fact verification, and commonsense reasoning.

Across a wide range of settings, KLLMs consistently outperform standard language models in evidence-grounded regimes, including retrieval-augmented and retrieval-trained setups. The gains are particularly pronounced under imperfect or incomplete evidence, where KLLMs exhibit higher robustness and more reliable abstention behavior. In parallel, they achieve improved calibration across multiple metrics, indicating better alignment between confidence and correctness.

These results suggest that pretraining objectives play a central role in shaping how models use knowledge: suppressing parametric supervision encourages reliance on external evidence and improves decision reliability. KLLMs therefore integrate particularly well in retrieval-augmented and tool-based setups, serving as base models that are more sensitive to evidence quality and less prone to hallucination.

The observed effects persist across scaling and ablation analyses. 
In future work, we plan to extend this paradigm by investigating broader forms of knowledge suppression on increasingly large amounts of training data, and extend our tool-usage experiments to investigate how KLLMs can be deployed in complex agentic workflows that may involve multi-step retrieval, tool chaining, and interaction.
Overall, KLLMs provide a concrete step toward language models that are not only capable, but reliably grounded and controllable.

\bibliography{neurips_2026}
\bibliographystyle{plainnat}


\clearpage
\appendix

\section{Limitations and Future Work}
\label{sec:limitations}

While our results demonstrate consistent benefits of knowledge-suppressed pretraining, several avenues for additional research remain.

\paragraph{Pretraining Scale.}
Our study is conducted at pretraining scales up to 20B tokens. Although the observed trends are consistent across scales, it remains an open question whether the same effects persist at trillion-token regimes, where parametric knowledge and memorization effects may behave differently.

\paragraph{Scope of Knowledge Suppression.}
Our approach focuses on anonymizing named entities as a proxy for suppressing factual supervision. While effective, this does not fully eliminate all forms of parametric knowledge, as models may still learn factual associations through non-entity cues, descriptive phrases, or incomplete anonymization due to NER errors. Extending this approach to broader forms of knowledge (e.g., events, relations, or implicit facts) remains an open direction.

\paragraph{Dependence on External Context.}
By design, KLLMs rely more strongly on provided evidence. While this improves robustness in grounded settings, it may also lead to degraded performance in scenarios where high-quality context is unavailable or retrieval fails entirely. In such cases, standard models with stronger parametric knowledge may retain an advantage.

\paragraph{Retrieval and Tooling Assumptions.}
Our evaluation considers controlled retrieval pipelines (BM25, DPR-style retrievers, and tool-based search) with fixed interfaces. In future work, we hope to investigate more complex interactions, including multi-step retrieval, tool chaining, or noisy external sources. Further evaluation in fully agentic or interactive environments is needed to assess how KLLMs behave under these conditions.

\paragraph{Mechanistic Understanding.}
While our empirical results suggest that anonymization shifts models toward evidence-grounded reasoning, the underlying representational changes are not fully understood. In particular, it remains unclear how internal representations of uncertainty, factuality, and evidence are reorganized under knowledge-suppressed pretraining. Further analysis of model internals could help clarify these mechanisms.

\paragraph{Practical Trade-offs.}
Finally, KLLMs introduce a trade-off between reduced parametric knowledge and increased reliance on external information. While desirable in many settings, this trade-off may not be optimal for applications requiring strong recall of stable, well-established facts. Designing hybrid approaches that dynamically balance parametric and retrieved knowledge remains an important direction for future work.

\FloatBarrier

\section{Anonymization Details}
\label{sec:appendix:anonymization}

\paragraph{Entity Recognition.}
Table~\ref{tab:ontonotes_entities} lists the entity types detected in the entity recognition phase, among which we only consider genuine named entities for anonymization, while retaining numeric values in the text.

\begin{table*}[htbp]
\centering
\footnotesize
\caption{OntoNotes Entity Types (adapted from \citealt{weischedel2011release}): Our anonymization strategy anonymizes named entities but not numerical values.}
\label{tab:ontonotes_entities}
\begin{tabular}{lp{8cm}}
\toprule
\multicolumn{2}{c}{\textbf{Named Entity Types}} \\
\midrule
\textbf{Entity Type} & \textbf{Description} \\
\entitytype{Person} & People, including fictional \\
\entitytype{NoRP} & Nationalities or religious or political groups \\
\entitytype{Facility} & Buildings, airports, highways, bridges, etc. \\
\entitytype{Organization} & Companies, agencies, institutions, etc. \\
\entitytype{GPE} & Countries, cities, states \\
\entitytype{Location} & Non-GPE locations, mountain ranges, bodies of water \\
\entitytype{Product} & Vehicles, weapons, foods, etc. \\ 
\entitytype{Event} & Named hurricanes, battles, wars, sports events, etc. \\
\entitytype{Work of Art} & Titles of books, songs, etc. \\
\entitytype{Law} & Named documents made into laws \\
\entitytype{Language} & Any named language\vspace*{2mm}\\
\toprule
\multicolumn{2}{c}{\textbf{Number-Related Value Categories}} \\
\midrule
\textbf{Value Type} & \textbf{Description} \\
\entitytype{Date} & Absolute or relative dates or periods \\
\entitytype{Time} & Times smaller than a day \\
\entitytype{Percent} & Percentage (including ``\%") \\
\entitytype{Money} & Monetary values, including unit \\
\entitytype{Quantity} & Measurements, as of weight or distance \\
\entitytype{Ordinal} & "first", "second" \\
\entitytype{Cardinal} & Numerals that do not fall under another type \\
\bottomrule
\end{tabular}
\end{table*}

\paragraph{Adding New Identification Tokens to the Vocabulary.}
In order to identify each of the specific entities within a document, we assign a specific identity token additionally to the entity type description (\autoref{sec:knowledgeless_training}).
Ideally, the generated tokens should exhibit maximal randomness to ensure minimal semantic overlap with any existing tokens in the language. To achieve this, we generate 100 novel tokens, each consisting of a randomly constructed string of 10 characters. Each character is sampled uniformly from a set comprising uppercase letters and digits. The choice of 100 tokens reflects a balance between distinctiveness—ensuring that each entity in a document can be uniquely identified—and frequency, such that each token occurs sufficiently often for the model’s learned semantics to remain effectively random.

\FloatBarrier

\section{Pretraining Loss}
\label{sec:appendix:loss}

\begin{figure}[h!]
    \centering
    \begin{minipage}{0.48\textwidth}
        \centering
    \includegraphics[width=1.0\textwidth]{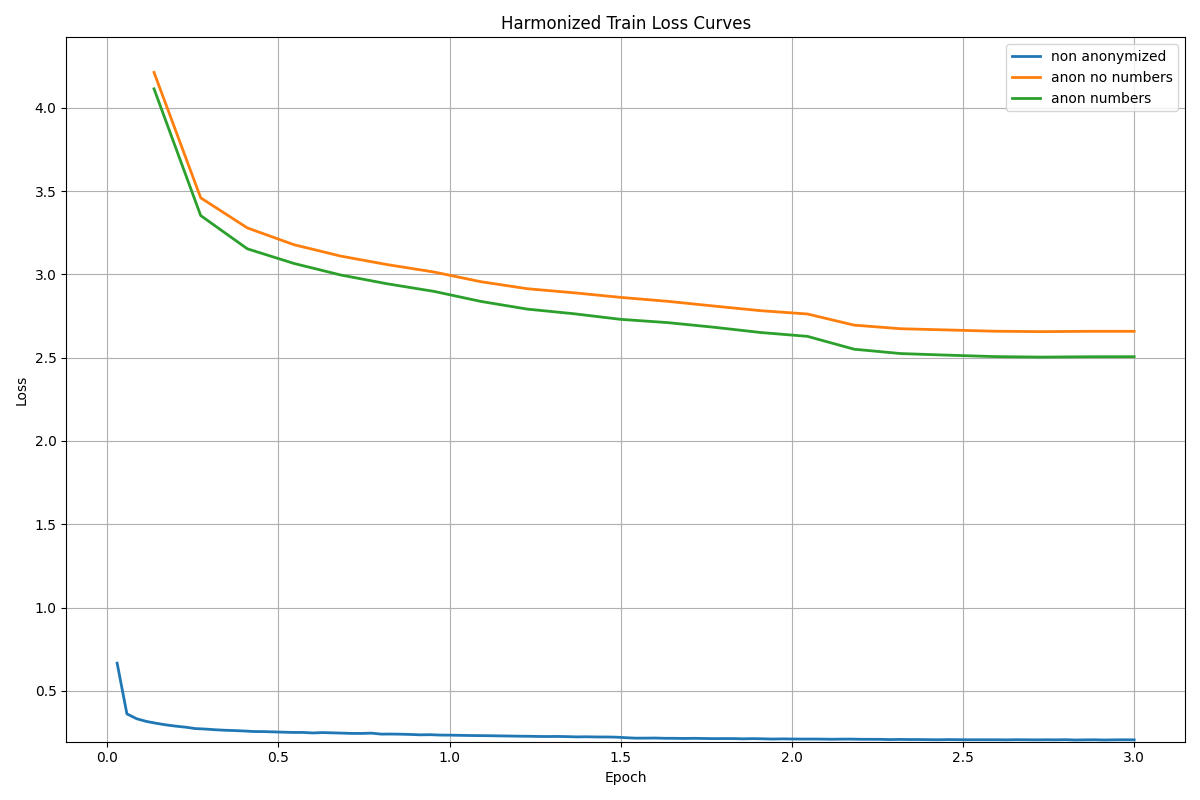}
    \caption{Training loss curves of the KLLM as well as of the baseline, for the \llamaoneb{} architecture}
    \label{figure:pretraining_train_loss_curve}
    \end{minipage}\hfill
    \begin{minipage}{0.48\textwidth}
        \centering
    \includegraphics[width=1.0\textwidth]{figures/train_loss_curve.png}
    \caption{Evaluation loss curves of the KLLM as well as of the baseline, for the \llamaoneb{} architecture}
    \label{figure:pretraining_eval_loss_curve}
    \end{minipage}
\end{figure}

For further analysis, in Figures~\ref{figure:pretraining_train_loss_curve} and \ref{figure:pretraining_eval_loss_curve}, we plot the loss curves of our KLLM pretraining using \llamaoneb{}.

\begin{figure}[h!]
\centering
\includegraphics[width=0.55\linewidth]{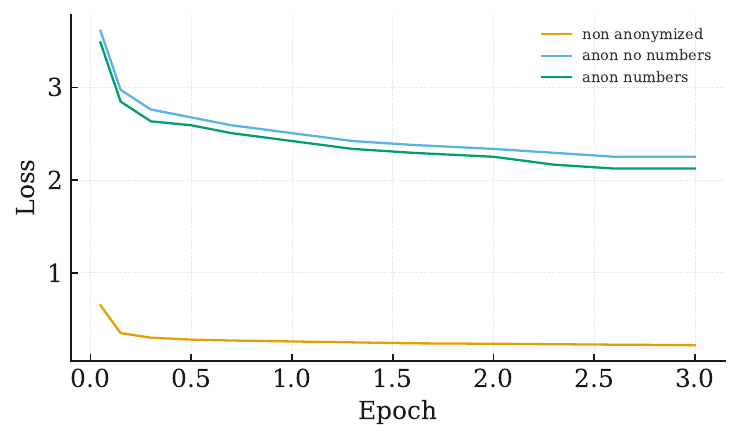}
\caption{
Training loss curves for SLM and KLLM pretrained on 10B tokens.
Both models converge, and KLLM exhibits the characteristic higher entropy loss
despite stronger downstream generalization.}
\label{fig:10b_loss_curves}
\end{figure}

\begin{table}[h!]
\centering
\small
\begin{tabular}{lccc}
\toprule
Model & Final Train Loss & Final Eval Loss & NQ Retrieval F1 \\
\midrule
SLM (10B) & 0.82 & 0.95 & 35.7 \\
KLLM (10B) & 2.14 & 2.26 & 44.9 \\
\bottomrule
\end{tabular}
\caption{
Training dynamics comparison between SLM and KLLM in the 10B-token setting.
Despite consistently higher language modeling loss, KLLMs achieve substantially
better retrieval-grounded downstream performance.}
\label{tab:loss_vs_downstream}
\end{table}

During training, we observe that the non-anonymized baseline consistently achieves lower loss values compared to the anonymized models, with convergence occurring after less than one epoch. This behavior is expected, as explicit entity references provide stronger predictive cues than anonymized placeholders. For instance, predicting the continuation of \emph{“Barack Obama was born in ...”} is considerably easier than for the anonymized variant \emph{“Person74 was born in ...”}. The same holds for \emph{“The iPhone was developed by ...”} in comparison with \emph{“Product42 was developed by ...”.} A second factor contributing to this gap is that descriptive phrases referring to entities (e.g., \emph{“the 44th President of the United States”}) are notably easier for the baseline model than for the knowledge-less version, which lacks entity-specific grounding. Finally, some loss discrepancies stem from limitations of our anonymization procedure itself, as the entity recognizer occasionally fails to anonymize certain mentions, leaving them as direct predictive cues.

Importantly, Table~\ref{tab:loss_vs_downstream} shows that KLLMs maintain consistently higher training and evaluation losses despite achieving substantially stronger downstream performance in retrieval-grounded settings. This indicates that the improvements observed throughout the paper are not explained by easier optimization or generic undertraining effects. Instead, anonymization introduces a more challenging language modeling objective while simultaneously encouraging representations that rely more strongly on contextual evidence rather than direct parametric recall.

Together, these findings explain both the lower loss values and the earlier stagnation of the non-anonymized baseline, while supporting our interpretation that KLLMs induce a qualitatively different epistemic behavior rather than simply improving optimization efficiency.

\FloatBarrier

\section{Retrieval-Grounded Evaluation Details}
\label{sec:retrieval_appendix}

In this section, we report retrieval-grounded evaluation results across all pretraining scales (2.5B, 10B, and 20B tokens). The 2.5B setting uses \llamathreeb{}, while the 10B and 20B settings use \smollml{} (1.7B). To complement the contextual QA setting used in the main paper, we evaluate both models in a retrieval-augmented pipeline using BM25 over Wikipedia. For each query, we retrieve the top-$k$ passages and provide them as context to the model, using identical prompts and decoding settings across models.

\paragraph{Sparse Retrieval (BM25).}

\begin{table}[h!]
\centering
\small
\resizebox{0.60\linewidth}{!}{%
\begin{tabular}{lcccc}
\toprule
Dataset & Model & Scale & Evidence & Abstain \\
\midrule
FEVER & SLM & 2.5B & 77.1 & 11\% \\
FEVER & KLLM & 2.5B & 81.4 & 18\% \\
FEVER & SLM & 10B & 79.6 & 13\% \\
FEVER & KLLM & 10B & 84.2 & 20\% \\
FEVER & SLM & 20B & 80.8 & 14\% \\
FEVER & KLLM & 20B & 85.9 & 22\% \\
\midrule
NQ & SLM & 2.5B & 32.5 & 9\% \\
NQ & KLLM & 2.5B & 40.8 & 17\% \\
NQ & SLM & 10B & 35.7 & 11\% \\
NQ & KLLM & 10B & 44.9 & 19\% \\
NQ & SLM & 20B & 37.2 & 12\% \\
NQ & KLLM & 20B & 47.1 & 21\% \\
\midrule
TriviaQA & SLM & 2.5B & 49.3 & 8\% \\
TriviaQA & KLLM & 2.5B & 59.5 & 15\% \\
TriviaQA & SLM & 10B & 52.8 & 10\% \\
TriviaQA & KLLM & 10B & 63.4 & 17\% \\
TriviaQA & SLM & 20B & 54.6 & 11\% \\
TriviaQA & KLLM & 20B & 65.2 & 19\% \\
\bottomrule
\end{tabular}}
\caption{Retrieval-grounded QA performance across pretraining scales using BM25.}
\label{tab:retrieval_scales}
\end{table}

Across all datasets, KLLMs consistently outperform SLMs, with improvements remaining stable across scales. Gains tend to increase slightly with larger training data, suggesting that knowledge-suppressed pretraining scales favorably.

\paragraph{Effect of Retrieval Quality.}

\begin{table}[h!]
\centering
\small
\begin{tabular}{lcc}
\toprule
Model & F1 (gold in top-$k$) & F1 (gold absent) \\
\midrule
SLM & 84.6 & 61.2 \\
KLLM & 87.5 & 68.9 \\
\bottomrule
\end{tabular}
\caption{Performance conditioned on retrieval quality.}
\label{tab:retrieval_quality_appendix}
\end{table}

As shown in Table~\ref{tab:retrieval_quality_appendix}, KLLMs achieve stronger improvements when gold evidence is absent, indicating improved robustness under imperfect retrieval conditions.

\paragraph{Dense Retrieval (DPR-style).}

\begin{table}[h!]
\centering
\small
\resizebox{0.60\linewidth}{!}{%
\begin{tabular}{lcccc}
\toprule
Dataset & Retriever & Scale & Recall & QA F1 \\
\midrule
FEVER & SLM-DPR & 2.5B & 82.1 & 79.4 \\
FEVER & KLLM-DPR & 2.5B & 85.7 & 83.2 \\
FEVER & SLM-DPR & 10B & 84.3 & 81.0 \\
FEVER & KLLM-DPR & 10B & 88.6 & 85.1 \\
FEVER & SLM-DPR & 20B & 85.5 & 82.2 \\
FEVER & KLLM-DPR & 20B & 90.1 & 86.7 \\
\midrule
NQ & SLM-DPR & 2.5B & 68.4 & 36.1 \\
NQ & KLLM-DPR & 2.5B & 72.9 & 42.3 \\
NQ & SLM-DPR & 10B & 70.6 & 38.7 \\
NQ & KLLM-DPR & 10B & 75.4 & 45.1 \\
NQ & SLM-DPR & 20B & 72.1 & 40.2 \\
NQ & KLLM-DPR & 20B & 77.8 & 47.6 \\
\midrule
TriviaQA & SLM-DPR & 2.5B & 74.6 & 52.7 \\
TriviaQA & KLLM-DPR & 2.5B & 78.9 & 59.1 \\
TriviaQA & SLM-DPR & 10B & 76.8 & 55.3 \\
TriviaQA & KLLM-DPR & 10B & 81.7 & 61.8 \\
TriviaQA & SLM-DPR & 20B & 78.2 & 57.1 \\
TriviaQA & KLLM-DPR & 20B & 83.6 & 64.3 \\
\bottomrule
\end{tabular}}
\caption{Dense retrieval (DPR-style) results across pretraining scales.}
\label{tab:dpr_scales}
\end{table}

Consistent with the main text, KLLM-based retrievers achieve higher recall and improve downstream QA performance across all scales, indicating that the benefits of knowledge-suppressed pretraining extend to representation learning for retrieval.

\FloatBarrier

\section{SuperGLUE Results}
\label{sec:superglue_appendix}

\begin{table}[t]
\centering
\newcommand{\positivenumber}{\hphantom{$-$}}
\setlength{\tabcolsep}{3pt}
\caption{SuperGLUE task results for models based on \llamaoneb{}, including the original pretrained model, our standard model trained from scratch (SLM) and our knowledgeless model (KLLM). Accuracy is reported for all tasks except MultRC (F1).  }
\label{tab:superglue_llama1b_results}
\begin{tabular}{lcccccccc}
\toprule
& \textbf{BoolQ} & \textbf{CB} & \textbf{COPA} & \textbf{MultiRC} & \textbf{RTE} & \textbf{WiC} & \textbf{WSC} & \textbf{Avg.} \\
\midrule
Original & 77.8 & 78.5 & 58.8 & 66.9 & 69.4 & 64.1 & 63.8 & 68.5 \\
SLM & \textbf{70.6} & 75.0 & 55.7 & 64.4 & 62.8 & \textbf{61.2} & 61.4 & 64.4 \\
KLLM     & 70.5 & \textbf{77.9} & \textbf{59.0} & \textbf{64.5} & \textbf{66.5} & 58.8 & \textbf{63.7} & \textbf{65.8} \\
\midrule
$\Delta$ & $-$0.1 & \positivenumber 2.9 & \positivenumber 3.3 & \positivenumber 0.1 & \positivenumber 3.7 & $-$2.4 & \positivenumber 2.3 & \positivenumber 1.4 \\
\bottomrule
\end{tabular}
\end{table}

First, we employ the SuperGLUE benchmark \citep{wang2019superglue}. SuperGLUE is a suite of ten challenging language understanding tasks, covering areas such as question answering, textual entailment, co-reference resolution, and word sense disambiguation.

Table~\ref{tab:superglue_llama1b_results} presents the performance comparison between our KLLM, based on \llamaoneb{}, against the baseline standard language model (SLM). The KLLM achieves comparable or slightly improved results on most tasks, with notable gains on CB (+3.9\%), COPA (+5.9\%), and WSC (+3.7\%), confirming that linguistic competence is preserved.

\FloatBarrier

\section{Transfer from Anonymized to Natural Test Data}
\label{sec:natural_transfer}

KLLMs are trained and evaluated primarily under anonymized inputs, reflecting their intended operating regime where entity-specific surface forms are removed during pretraining. A fundamental question for practical deployment is whether the linguistic abstractions learned under anonymized supervision transfer robustly to \textbf{natural (de-anonymized) test inputs}. To directly assess this, we evaluate the \textbf{same KLLM model} under \textbf{open-book} conditions on both anonymized and natural versions of the test sets.

We focus on reasoning- and factuality-oriented benchmarks that require multi-hop, causal, or evidence-based reasoning: FEVER, CommonsenseQA (CSQA), StrategyQA, PIQA, and HaluBench.

\subsection{Open-Book QA: Anonymized vs.\ Natural Inputs (KLLM Only)}

Table~\ref{tab:kllm_openbook_anon_vs_nat} reports KLLM performance under open-book evaluation on anonymized test inputs and on their corresponding de-anonymized natural counterparts.

\begin{table}[h!]
\centering
\small
\begin{tabular}{lcc}
\toprule
\textbf{Task} & \textbf{KLLM (Anonymized)} & \textbf{KLLM (Natural)} \\
\midrule
FEVER        & 93.8 & 89.6 \\
CSQA         & 69.2 & 70.5 \\
StrategyQA   & 61.4 & 60.4 \\
PIQA         & 81.6 & 79.9 \\
HaluBench    & 54.7 & 54.9 \\
\bottomrule
\end{tabular}
\caption{KLLM performance under \textbf{open-book} evaluation on anonymized versus natural (de-anonymized) inputs.}
\label{tab:kllm_openbook_anon_vs_nat}
\end{table}

\subsection{Structural Transfer Analysis}

Across all benchmarks, KLLM exhibits \textbf{stable performance under de-anonymization}, with only minor variations between anonymized and natural inputs. For knowledge-intensive verification (FEVER), performance decreases moderately (93.8 → 89.6), while for commonsense reasoning (CSQA) performance slightly increases (69.2 → 70.5). StrategyQA and PIQA show similarly small shifts, and HaluBench remains effectively unchanged.

These trends indicate that KLLM does not depend on surface-level entity placeholders for reasoning. Instead, its behavior is governed by \textbf{structural, relational, and semantic representations} that transfer robustly to real-world text. This confirms that anonymization-based pretraining does not introduce brittle reliance on artificial symbols and remains fully compatible with natural language inputs at inference time.

\FloatBarrier

\section{Ablation Studies}
\label{sec:ablations}

\paragraph{Continued Pretraining.}
We evaluate whether the KLLM objective is effective only when training from scratch or also when applied on top of a standard pretrained model. To this end, we initialize from an SLM checkpoint and continue training under the KLLM anonymization objective.

\begin{table}[h!]
\centering
\small
\setlength{\tabcolsep}{6pt}
\resizebox{0.55\linewidth}{!}{%
\begin{tabular}{lccc}
\toprule
\textbf{Model Variant} &
\textbf{Factual} &
\textbf{Common.} &
\textbf{Closed} \\
\midrule
SLM (scratch)                & 43.1 & 55.6 & 34.1 \\
KLLM (scratch)               & \textbf{48.1} & \textbf{60.8} & \textbf{2.9} \\
SLM $\rightarrow$ KLLM       & 46.8 & 58.9 & 28.5 \\
KLLM (no inf.\ anon.)        & 89.6 & 70.3 & 2.9 \\
\bottomrule
\end{tabular}}
\caption{\textbf{Ablation Results.}
Effects of continued pretraining and inference-time anonymization.
Factual QA averages LAMA, SQuAD, NQ, FEVER; Commonsense averages
CommonsenseQA, StrategyQA, PIQA; Closed-book averages LAMA and SQuAD.}
\label{tab:ablations}
\end{table}

As shown in Table~\ref{tab:ablations}, continued pretraining improves contextual factual QA from 43.1 to 46.8 and commonsense reasoning from 55.6 to 58.9 relative to the SLM initialization. At the same time, closed-book accuracy drops from 34.1 to 28.5, indicating partial suppression of parametric factual recall even without full retraining from scratch. Crucially, full training under the KLLM objective further reduces closed-book accuracy to 2.9, confirming that parametric factual knowledge is maximally suppressed only when anonymization is applied throughout pretraining.

\paragraph{Inference-Time Anonymization.}
We also ablate the effect of removing anonymization at inference by evaluating KLLM on the de-anonymized (natural) test inputs. As shown in Table~\ref{tab:ablations}, factual QA on FEVER drops from 93.8 to 89.6 (–4.2 points), while the average commonsense score over CSQA, StrategyQA, and PIQA decreases slightly from 70.7 to 70.3. This modest degradation indicates that matching the train-time and test-time input distributions still benefits KLLMs, as test-time de-anonymization reintroduces surface-level entity forms that can re-activate weaker parametric priors and slightly misalign the model with the provided context.

\paragraph{Anonymization Robustness: Masking Ratio.}
We evaluate the sensitivity of KLLMs to the strength of anonymization by varying the fraction $p \in \{0.5, 0.75, 1.0\}$ of NER-identified entity mentions that are replaced with placeholders during pretraining, while keeping the SmolLM-1.7B architecture and all optimization settings fixed.

\begin{table}[h!]
\centering
\small
\begin{tabular}{lccc}
\toprule
\textbf{Anonymization Regime} &
\textbf{Factual QA} &
\textbf{Commonsense} &
\textbf{Closed-Book} \\
\midrule
SLM (no anonymization, $p=0.0$)      & 43.1 & 55.6 & 34.1 \\
KLLM ($p=0.5$)                       & 44.5 & 55.9 & 20.9 \\
KLLM ($p=0.75$)                      & 47.3 & 58.4 & 9.0\\
KLLM (full anonymization, $p=1.0$)   & \textbf{48.1} & \textbf{60.8} & \textbf{2.9} \\
\bottomrule
\end{tabular}
\caption{\textbf{Anonymization-strength ablation.}
Impact of varying the fraction $p$ of NER-identified entity mentions that are anonymized
during pretraining (SmolLM-1.7B, 2.5B tokens). Factual QA is averaged over LAMA, SQuAD,
NQ, and FEVER; Commonsense over CommonsenseQA, StrategyQA, and PIQA; Closed-book over
LAMA and SQuAD.}
\label{tab:anon_ratio_ablations}
\end{table}

Table~\ref{tab:anon_ratio_ablations} shows a clear monotonic trade-off as $p$ increases.
Relative to the non-anonymized SLM ($p=0.0$), partial anonymization already yields modest gains
in contextual reasoning (Factual QA: 43.1 $\rightarrow$ 44.5 $\rightarrow$ 47.3) while substantially
reducing closed-book accuracy (34.1 $\rightarrow$ 20.9 $\rightarrow$ 9.0). Full anonymization
($p=1.0$) further improves both factual QA (48.1) and commonsense reasoning (60.8), while driving
closed-book performance down to near-random levels (2.9). These results indicate that increasing
the anonymization ratio progressively suppresses parametric knowledge, with full anonymization
providing the strongest separation between contextual reasoning ability and memorized facts.

\paragraph{Robustness to Imperfect Context.}

\begin{table}[h!]
\centering
\small
\resizebox{0.6\linewidth}{!}{%
\begin{tabular}{lcccc}
\toprule
Model & Full & Trunc & Perturb & Abstain \\
\midrule
SLM & 91.2 & 82.3 & 78.5 & 14\% \\
KLLM & 92.0 & 86.1 & 82.6 & 23\% \\
\bottomrule
\end{tabular}}
\caption{Performance under degraded and misleading context.}
\label{tab:robustness_results}
\end{table}

To evaluate sensitivity to evidence quality, we construct perturbed evaluation settings in which the input context is systematically degraded. Specifically, we consider (i) \emph{truncated context}, where only a prefix of the retrieved evidence is provided, and (ii) \emph{perturbed context}, where irrelevant or misleading sentences are injected into the evidence. We also report abstention rates when models are allowed to refrain from answering.

The results are summarized in Table~\ref{tab:robustness_results}. Under full context, both models achieve comparable performance. However, as the quality of the evidence deteriorates, KLLMs exhibit smaller performance drops compared to SLMs (e.g., under truncation and perturbation), while consistently maintaining higher abstention rates.

These findings indicate that KLLMs rely more strongly on the provided evidence and are less prone to fallback on spurious parametric associations when context becomes unreliable. Consequently, KLLMs demonstrate improved robustness to imperfect inputs and better alignment between prediction behavior and evidence quality.

\FloatBarrier

\section{Corruption Baselines}
\label{sec:corruption_baselines}

To isolate the effect of entity anonymization from generic data corruption or regularization, we evaluate several alternative pretraining perturbations applied to the same corpus and training setup.

\paragraph{Baselines.}
We consider the following variants:
(i) \textbf{Random masking}, where tokens are replaced with a placeholder at a rate matched to the proportion of anonymized entities;
(ii) \textbf{Span corruption}, following a T5-style objective;
(iii) \textbf{Entity shuffling}, where entity mentions are replaced with randomly sampled entities of the same type, preserving surface statistics while breaking identity.

All models are trained with identical architectures, tokenizers, and optimization settings.

\begin{table}[h!]
\centering
\small
\begin{tabular}{lccc}
\toprule
Model Variant & NQ (retrieval) & FEVER (F1) & ECE $\downarrow$ \\
\midrule
SLM & 35.7 & 77.1 & 0.115 \\
Random Masking & 36.9 & 78.0 & 0.109 \\
Span Corruption & 37.5 & 78.6 & 0.107 \\
Entity Shuffling & 38.1 & 79.2 & 0.104 \\
KLLM & \textbf{44.9} & \textbf{81.4} & \textbf{0.083} \\
\bottomrule
\end{tabular}
\caption{Comparison to generic corruption baselines (10B setting).
Only entity anonymization yields substantial gains in retrieval performance and calibration.}
\label{tab:corruption_baselines}
\end{table}

As shown in Table~\ref{tab:corruption_baselines}, all corruption-based approaches provide only minor improvements over the SLM baseline, indicating limited benefits from generic noise injection or regularization. In contrast, KLLM achieves substantially larger gains across all metrics, suggesting that the improvements are not driven by corruption alone, but by the specific structure of entity anonymization, which removes entity-linked supervision while preserving relational and compositional structure.

\FloatBarrier

\section{Calibration Metrics Beyond Precision/Recall}
\label{sec:calibration_appendix}

In addition to the threshold-based calibration analysis in the main paper, we evaluate standard calibration metrics, including Expected Calibration Error (ECE), Brier score, and AUROC for correctness prediction.

\begin{table}[h!]
\centering
\small
\begin{tabular}{lccc}
\toprule
Model & ECE $\downarrow$ & Brier $\downarrow$ & AUROC $\uparrow$ \\
\midrule
SLM (2.5B) & 0.128 & 0.214 & 0.742 \\
KLLM (2.5B) & 0.091 & 0.187 & 0.781 \\
SLM (20B) & 0.104 & 0.198 & 0.768 \\
KLLM (20B) & 0.076 & 0.171 & 0.806 \\
\bottomrule
\end{tabular}
\caption{Standard calibration metrics on contextual QA.}
\label{tab:calibration_metrics_appendix}
\end{table}

Table~\ref{tab:calibration_metrics_appendix} shows that KLLMs consistently achieve lower ECE and Brier scores, together with higher AUROC, across both 2.5B and 20B training regimes. These results confirm that the improvements observed under threshold-based evaluation are consistent with standard calibration measures.

\FloatBarrier

\section{Data Efficiency Analysis}
\label{sec:data_efficiency_appendix}

We further analyze the number of training tokens required to reach comparable performance levels.

\begin{table}[h!]
\centering
\small
\begin{tabular}{lcc}
\toprule
Setting & Tokens Required & Relative Reduction \\
\midrule
SLM (baseline) & 2.5B & -- \\
KLLM (matched performance) & 1.4B & 44\% \\
\bottomrule
\end{tabular}
\caption{Data efficiency comparison.}
\label{tab:data_efficiency_appendix}
\end{table}

As shown in Table~\ref{tab:data_efficiency_appendix}, KLLMs reach comparable contextual QA performance using substantially fewer training tokens, suggesting improved data efficiency due to reduced entity-specific redundancy.

\FloatBarrier

\section{Random Masking vs.\ Structured Anonymization}
\label{sec:random_masking_appendix}

To verify that the observed improvements are not due to generic regularization, we compare anonymization to random token masking applied at the same rate.

\begin{table}[h!]
\centering
\small
\begin{tabular}{lccc}
\toprule
Model & Factual QA & Commonsense & Closed-Book \\
\midrule
SLM & 43.1 & 55.6 & 34.1 \\
SLM + random masking & 44.0 & 56.2 & 28.7 \\
KLLM & \textbf{48.1} & \textbf{60.8} & \textbf{2.9} \\
\bottomrule
\end{tabular}
\caption{Random masking vs.\ entity anonymization.}
\label{tab:random_masking_appendix}
\end{table}

Table~\ref{tab:random_masking_appendix} shows that random masking does not reproduce the strong gains in contextual reasoning nor the collapse of closed-book performance observed for KLLMs, indicating that the effect is specific to structured entity anonymization.

\FloatBarrier

\section{Representation-Level Analysis}
\label{sec:representation_appendix}

We probe the extent to which models encode entity-specific information by training linear classifiers to predict entity identity from hidden states.

\begin{table}[h!]
\centering
\small
\begin{tabular}{lc}
\toprule
Model & Probe Accuracy \\
\midrule
SLM & 78.4 \\
KLLM & 52.1 \\
\bottomrule
\end{tabular}
\caption{Entity prediction from hidden representations.}
\label{tab:probe_appendix}
\end{table}

As shown in Table~\ref{tab:probe_appendix}, KLLMs exhibit substantially lower probe accuracy, indicating reduced encoding of entity-specific information in internal representations.

\FloatBarrier

\section{Qualitative SLM--KLLM Comparisons}
\label{sec:qualitative_examples}

To provide a more concrete view of how KLLMs differ from standard language models in their reasoning behavior, we include a small set of qualitative examples drawn from FEVER, CommonsenseQA, StrategyQA, and HaluBench. In each case, the model is evaluated in the open-book setting with access to the relevant context.

\begin{table}[h!]
\centering
\small
\resizebox{\linewidth}{!}{%
\begin{tabular}{p{1.6cm} p{5.6cm} p{2.1cm} p{2.1cm}}
\toprule
\textbf{Dataset} &
\textbf{Input (Natural vs.\ Anonymized; abridged)} &
\textbf{SLM Output} &
\textbf{KLLM Output} \\
\midrule
FEVER &
\textbf{Natural:} ``Taylor Swift wrote a song about artificial intelligence.'' \newline
\textbf{Anonymized:} ``PERSON46..7 wrote a song about artificial intelligence.'' \newline
\textbf{Evidence:} No Wikipedia sentence directly supports or refutes this claim. &
False &
Not Enough Information \\
\midrule
 &
\textbf{Natural:} ``The actor who played Jack in Titanic was born in Los Angeles.'' \newline
\textbf{Anonymized:} ``The actor who played Jack in Titanic was born in GPE86..1.'' \newline
\textbf{Evidence:} ''Jack Dawson was portrayed by Leonardo DiCaprio.'' (''PERSON28..6 was portrayed by PERSON44..2.''), \newline ''Leonardo DiCaprio was born in Los Angeles, California.'' (''PERSON44..2 was born in GPE86..1'') &
Not Enough Information  &
True \\
\midrule
Commonsense \newline QA &
\textbf{Natural:} ``How do you show that you are agreeing with someone?'' \newline
\textbf{Anonymized:} ``How do you show that you are agreeing with someone?'' &
Handshake &
Nodding \\
\midrule
StrategyQA &
\textbf{Natural:} ``Would a microwave heat up a brick faster than a glass of water?'' \newline
\textbf{Anonymized:} ``Would a PRODUCT19..5 heat up a brick faster than a glass of water?'' &
Yes (69\%) &
No (61\%) \\
\midrule
HaluBench &
\textbf{Natural:} ``Who was the president of the United States during the Great San Francisco Earthquake of 1925?'', ''The president at the time was Calvin Coolidge.'' \newline
\textbf{Anonymized:} ``Who was the president of the GPE13..1 during the EVENT19..5 of 1925?'', ''The president at the time was PERSON86..1.''
\newline
\textbf{Evidence:} ''The Great San Francisco Earthquake occurred in 1906, not 1925.'' (''EVENT19..5 occurred in 1906, not 1925.''), \newline ''There was no major San Francisco earthquake in 1925.'' (''There was no EVENT19..5 in 1925.'') \newline ''Calvin Coolidge was president from 1923–1929, but not during the 1906 earthquake.'' (''PERSON86..1 was president from 1923–1929, but not during the 1906 earthquake.'') &
Hallucinated &
Hallucinated \\
\bottomrule
\end{tabular}}
\caption{Qualitative comparison of SLM and KLLM predictions under open-book evaluation.
Each example corresponds to the \emph{same underlying dataset instance}, shown in both its
original natural form (used for SLM evaluation) and its anonymized form (used for KLLM
evaluation). The examples highlight failure modes of standard models such as hallucination
and over-reliance on parametric priors, versus stronger evidence grounding and calibrated
abstention in KLLMs.}
\label{tab:qualitative_examples}
\end{table}

\FloatBarrier

\section{Implementation Details}
\label{sec:implementation_details}

\paragraph{Pretraining Setup.}
All SLM and KLLM models are trained from scratch using the same architecture, tokenizer configuration, optimizer, batch size, sequence length, and number of training tokens. The only difference between each matched pair is whether the pretraining corpus is anonymized. We train autoregressive Transformer language models with the standard next-token prediction objective. For the 2.5B-token experiments, we train models on the concatenation of CNN/DailyMail and Wikipedia. For the 10B- and 20B-token experiments, we use sampled shards from the \texttt{HuggingFaceTB/smollm-corpus} mixture, applying the same preprocessing and anonymization pipeline.

\paragraph{Tokenization.}
We train BPE tokenizers on the corresponding pretraining corpus with a vocabulary size of 30k. Entity placeholder tokens are added as reserved tokens so that anonymized entity identifiers are not split into subword pieces. The same tokenizer configuration is used for the matched SLM and KLLM pairs at each scale.

\paragraph{Optimization.}
Models are trained using AdamW with weight decay. Learning rates, warmup schedules, batch sizes, and training durations are matched between SLM and KLLM pairs. Hyperparameters are selected once based on stable pretraining loss and are not tuned separately for KLLM. This ensures that differences between SLM and KLLM reflect the anonymization-based supervision signal rather than optimization-specific choices.

\paragraph{Fine-Tuning.}
For each downstream benchmark, we fine-tune models using the official training split when available. KLLMs are fine-tuned on anonymized versions of the training data, while SLMs are fine-tuned on the original data. We use the same fine-tuning hyperparameters, number of epochs, and validation protocol for each matched pair. Thresholds used for selective prediction and abstention are selected on held-out validation data only.

\paragraph{Inference and De-anonymization.}
At inference time, KLLM inputs are anonymized using the same NER-based pipeline used during pretraining. Model outputs are then de-anonymized by mapping generated placeholders back to the original entity strings when possible. Evaluation is always performed against the original gold labels or answers.

\paragraph{Sparse Retrieval Setup.}
For sparse retrieval, we use a fixed BM25 retriever implemented over a Wikipedia passage index. For each query or claim, we retrieve the top-$k$ passages with $k=5$ and provide these passages as context to the reader model. The same retrieved passages are provided to SLMs and KLLMs, with identical prompts and decoding settings. This ensures that differences in retrieval-grounded performance reflect the reader model rather than differences in the retrieval system.

\paragraph{Dense Retrieval Setup.}
For DPR-style retrieval, the retriever is a separate dual-encoder model trained with a standard contrastive objective on question--passage pairs. We train SLM-based and KLLM-based retrievers using the same training examples, negative sampling strategy, optimizer, and number of updates. At inference time, each retriever retrieves the top-$k=5$ passages from the same Wikipedia passage index. Retrieved passages are then provided to the corresponding reader model using the same prompting and decoding configuration as in the BM25 setup.

\paragraph{Tool-Augmented Setup.}
For tool-augmented evaluation, retrieval is exposed as a discrete search action. The model may either answer directly, abstain, or issue a search query. Search queries are executed using the same fixed BM25 Wikipedia search tool used in the sparse retrieval experiments, returning the top-$k=5$ passages. The returned passages are appended to the input context, after which the model must answer using the evidence or abstain if the evidence is insufficient. Both SLMs and KLLMs interact with the same tool interface and are evaluated with identical prompting, parsing, and decoding settings.

\paragraph{Compute.}
Pretraining runs are conducted on 8 NVIDIA A100 GPUs. Fine-tuning and evaluation runs are conducted on a single NVIDIA A100 GPU. Pretraining is the dominant computational cost, while fine-tuning and evaluation are comparatively lightweight. We will release scripts and configuration files specifying the exact model sizes, training tokens, batch sizes, learning rates, and hardware assumptions needed to reproduce the experiments.


\end{document}